\documentclass[a4paper,10pt,conference]{IEEEtran}
\IEEEoverridecommandlockouts
\usepackage{cite}
\usepackage{amsmath,amssymb,amsfonts}
\usepackage{algorithmic}
\usepackage{graphicx}
\usepackage{textcomp}
\usepackage{xcolor}
\usepackage{nidanfloat}
\usepackage{algorithm,algorithmic}
\usepackage{ulem}

\def\BibTeX{{\rm B\kern-.05em{\sc i\kern-.025em b}\kern-.08em
    T\kern-.1667em\lower.7ex\hbox{E}\kern-.125emX}}
\begin{document}

\title{A universal detector of CNN-generated images using properties of checkerboard artifacts in the frequency domain\\
}

\author{\IEEEauthorblockN{1\textsuperscript{st} Miki Tanaka}
\IEEEauthorblockA{\textit{Tokyo Metropoliltan University} \\
Tokyo, Japan \\
tanaka-miki@ed.tmu.ac.jp}
\and
\IEEEauthorblockN{2\textsuperscript{nd} Sayaka Shiota}
\IEEEauthorblockA{\textit{Tokyo Metropoliltan University} \\
Tokyo, Japan \\
sayaka@tmu.ac.jp}
\and
\IEEEauthorblockN{3\textsuperscript{rd} Hitoshi Kiya}
\IEEEauthorblockA{\textit{Tokyo Metropoliltan University} \\
Tokyo, Japan \\
kiya@tmu.ac.jp}
}

\maketitle

\begin{abstract}
We propose a novel universal detector for detecting images generated by using CNNs. In this paper, properties of checkerboard artifacts in CNN-generated images are considered, 
and the spectrum of images is enhanced in accordance with the properties. 
Next, a classifier is trained by using the enhanced spectrums to judge a query image to be a CNN-generated ones or not. In addition, an ensemble of the proposed detector with emphasized spectrums and a conventional detector is proposed to improve the performance of these methods.
In an experiment, the proposed ensemble is demonstrated to outperform a state-of-the-art method under some conditions.
\end{abstract}


\section{Introduction}
Convolutional neural networks (CNNs) have led to major breakthroughs in a wide range of applications. In contrast, they have generated new concerns and problems. Recent rapid advances in deep image synthesis techniques, such as generative adversarial networks (GANs) have easily generated fake images, so detecting manipulated images has become an urgent issue\cite{overview,robusthash_for_fake}.

To overcome this issues, we propose a universal detector of images generated by using CNNs in this paper. Recently, CNN-generated images were investigated to include a trace of checkerboard artifacts \cite{chekerboard1,checkerboard3,checkerboard2,chekerboard4}, although the trance is weak in general. We focus on a trace of checkerboard artifacts in the frequency domain to detect CNN-generated images. 

In the proposed method, the spectrum of images is enhanced in accordance with properties of checkerboard artifacts, and a classifier is trained by using the enhanced spectrums. In addition, an ensemble of the trained detector and a state-of-the-art detector\cite{CNN-gimg} is proposed for improving the performance of the detectors. In an experiment, the proposed ensemble is demonstrated to outperform the state-of-the-art one under the use of 11 models.

\section{Related work}
\subsection{Generator model}
A variety of generator models have been proposed for image generation and image-to-image translation. Models based on variational autoencoder (VAE) or GAN are typical image generator models\cite{originalGAN,VAE}. Autoencoders including VAE translate an image into latent variable $z$ by using an encoder, and generate an image from $z$ by using a decoder. Since $z$ has the standard normal distribution, VAE can generate images from a noise having the standard normal distribution by the decoder. Deepfakes\cite{deepfake_github} with VAE become a major threat to the international community.

GAN models estimate a generative model and a discriminative model via an adversarial process. 
PGGAN\cite{PGGAN}, BigGAN\cite{BigGAN}, StyleGAN\cite{StyleGAN} and StyleGAN2\cite{StyleGAN2} generate high resolution images from a random noise vector. In addition, CycleGAN\cite{cyclegan} and StarGAN\cite{stargan} translate an image from a source domain to a target domain, e.g. changing apples to oranges. GauGAN\cite{GauGAN} were proposed to generate an image from an input semantic layout.

\subsection{Detecting CNN-generated images}
The first approach for detecting CNN-generated images was inspired by photo-response non-uniformity noise (PRNU) that was used for discriminating camera devices\cite{ganfingerprint,ganfingerprint2}. This approach enables us to discriminate GAN-generated images from fingerprints caused by GAN, and assume that the same GAN models are used for training and testing images. 

To universally detect CNN-generated images even when images are not ones generated from a model used for training the detector, a universal detector was proposed, where it was trained by using images generated ony from AutoGAN\cite{AutoGAN}. In this work\cite{AutoGAN}, the use of the frequency domain was demonstrated to improve the performance of the detector.
In contrast, a universal detector by training only one specific GAN (PGGAN) was proposed\cite{CNN-gimg}, where RGB images are directly used for training a detector. This method was shown to detect not only PGGAN but also other generator models that were not used for training the detector.

\section{Proposed detector}
\subsection{Detector with enhanced spectrum}
\begin{figure}[h]
    \centerline{
    \includegraphics[width=8cm]{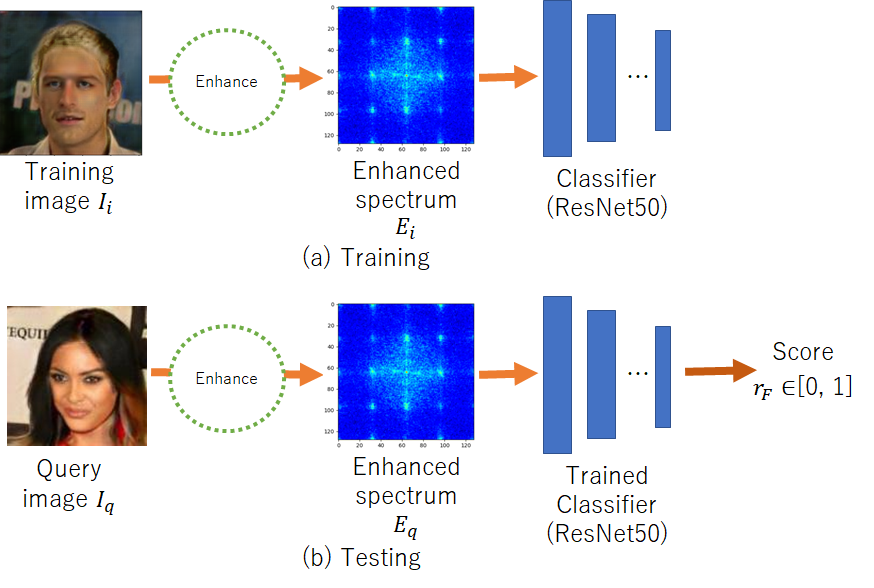}}
    \caption{Overview of proposed detector (single)}
    \label{fig:spfakedetection}
\end{figure}

Figure \ref{fig:spfakedetection} shows an overview of the proposed detector with enhanced spectrums. For training a classifier, a novel enhancement method for clearly showing checkerboard artifacts included in CNN-generated images is applied to training image $I_i$. Enhanced spectrum $E_i$ calculated from $I_i$ are used for training the classifier. For testing, a query image $I_Q$ is enhanced as well as for training, the spectrum $F_Q$ calculated from $I_Q$ is inputted to the trained classifier to judge it to be a CNN-generated one or not in accordance with an outputted probability score $r_F \in [0,1]$. 

\begin{figure}[h]
    \centerline{
    \includegraphics[width=8cm]{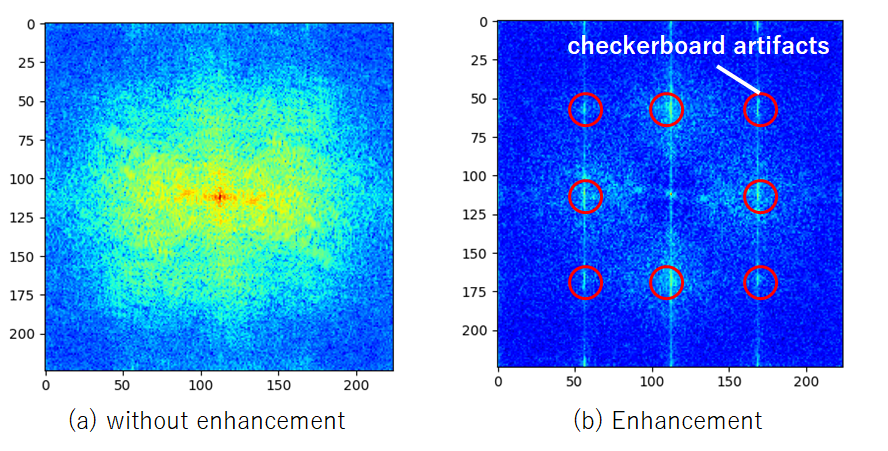}}
    \caption{Example of enhanced spectrum}
    \label{fig:enhanced}
\end{figure}

Enhanced spectrum $F_i$ is calculated as below. 
\begin{enumerate}
    \item A median filter with a size of $5 \times 5$ is appliled to $I_i$, and the difference between $I_i$ and $I'_i$, $I^D_i = I_i-I'_i$ is calculated, where $I'_i$ is an output image from the median filter.
    \item Cropping $I^D_i$ into $L$ rectangles with a size of $N \times N$ at random positions to generate $L$ images with a size of $N \times N$, $I^1_i, \dots, I^L_i$.
    \item Applying $N \times N$-DFT to $I^1_i, \dots, I^L_i$ to obtain their spectrums $F^1_i, \dots, F^L_i$.
    \item Computing enhanced spectrum $E_i$ as.
\end{enumerate}
\begin{eqnarray}
    E_i = \sum_{n=1}^L \log_{10}|F^n_i|
    \label{eq:F_E}
\end{eqnarray}
Similarly, enhanced spectrum $E_q$ is calculated by using query $I_q$ (see Fig.\ref{fig:enhanced}).

From properties of checkerboard artifacts, CNN-generated images are confirmed to have the influence of checkerboard artifacts at the same positions in the frequency domain for all cropped images. Accordingly, averaging spectrums as shown in the procedure can enhance the features that CNN-generated images have.

\subsection{Ensemble of RGB image and enhanced spectrum}
A state-of-the-art method for detecting CNN-generated images \cite{CNN-gimg} is directly carried out with RGB images. In this paper, we also propose an ensemble of this conventional detector and the detector with enhanced spectrums. 

\begin{figure}[ht]
    \centerline{
    \includegraphics[width=8cm]{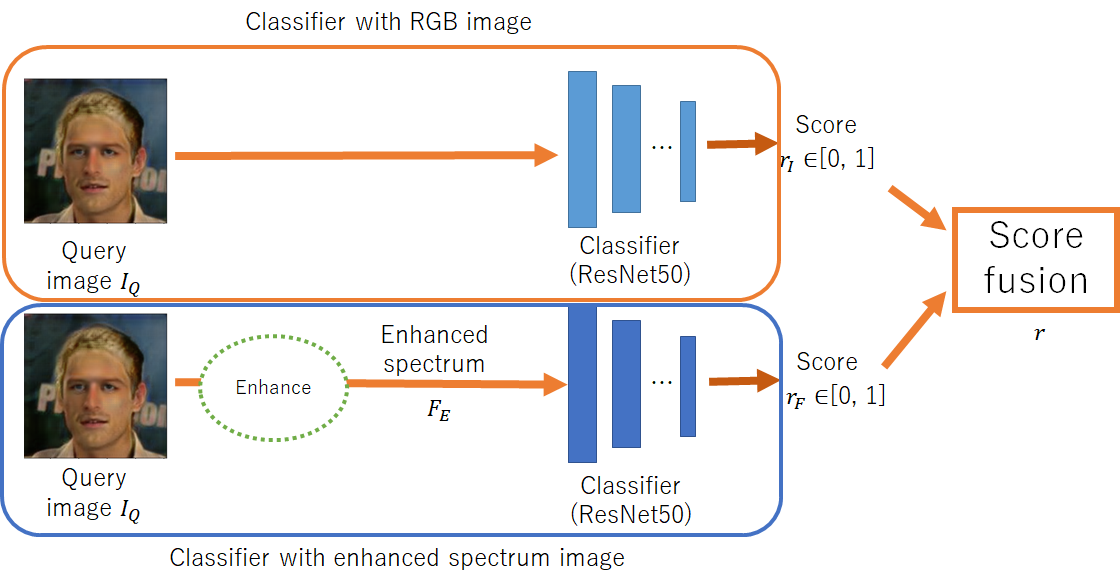}}
    \caption{Proposed detector (ensemble)}
    \label{fig:2streamfakedetection}
\end{figure}

An overview of the ensemble is shown in Fig. \ref{fig:2streamfakedetection}. Final probability score $r$ is calculated from $r_I$ and $F_F$ in accordance with Algorithm 1, where $r_I$ and $r_F$ are scores from the detector with enhanced spectrums and the detector with RGB images, respectively. Each detector has their own strengths and weaknesses, so this ensemble is expected to improve the performance of each detector. 

\begin{algorithm}[h]
    \caption{the ensemble algorithm}
    \begin{algorithmic}[1]
     \IF {($|r_I-0.5|>|r_F-0.5|$)}
     \STATE $r = r_I $ 
     \ENDIF
     \IF {($|r_I-0.5|<|r_F-0.5|$)}
     \STATE $r = r_F $ 
     \ENDIF
    \end{algorithmic} 
\end{algorithm}

\section{Experiment results}
\subsection{Experiment setup}
In this experiment, the dataset prepared by Wang et al. \cite{CNN-gimg} was used, where the training dataset consist of 720K images generated by using PGGAN \cite{PGGAN}, and test datasets consisted of 4K images in which 2K images are generated by using 11 models as CNN-generated images, and the others are images captured by cameras. 

\color{black}
The performance of detectors was evaluated by using  F-score and average precision (AP). F-score is given by,
\begin{eqnarray}
    \text{F-score} = \frac{2RP}{R+P,}
\end{eqnarray}
where $P$ and $R$ are the precision and recall at a selected threshold $th$, and $th = 0.5$ was selected for this evaluation. 
AP is also computed by summarizing a precision-recall curve as the weighted mean of precisions achieved at each threshold:
\begin{eqnarray}
    \text{AP} = \Sigma_j (R_j - R_{j-1})P_j\ , 
\end{eqnarray}
where $P_j$ and $R_j$ are the precision and recall at the $j_{th}$ threshold, and an input image is judged to be a CNN-generated image, when the probability score $r=[0,1]$ of the input image is higher than $th$. 

\color{black}

\subsection{Experiment results}
Table \ref{tb:result_f} and \ref{tb:result_ap} show AP and F-score values under the use of test images from each model.
\color{black} Table \ref{tb:result_f} shows \color{black} the proposed method (single) outperformed Wang's method under a number of models \color{black} that had similar network structures to PGGAN. In addition, \color{black} the proposed method (ensemble) outperformed Wang's method for almost all models. 
Wang's method and the proposed method (single) have their own strengths and weaknesses for detecting CNN-generated images, respectively. In contrast, the ensemble enables us to adopt only strong points of each method. Table \ref{tb:result_ap} also shows mean values calculated from AP values of 11 models. From the table, the proposed detector with enhanced spectrums was confirmed to improve the accuracy of Wang’s method, although the models were trained only by using images from PGGAN.
The difference between Table \ref{tb:result_f} and Table \ref{tb:result_ap} was caused due to the difference in the selection of threshold values.

\begin{table*}[t]
    \caption{Experiment result under 11 models (F-score)}
    \label{tb:result_f}
    \begin{center}
        \scalebox{0.80}{
    \begin{tabular}{c|c||c|c|c|c|c|c|c|c|c|c|c}
    Methods& Input image & PGGAN& StyleGAN & StyleGAN2 & BigGAN & CycleGAN & StarGAN & GauGAN & CRN & IMLE & SAN & Deepfake\\ \hline
    \hline
    Wang's method\cite{CNN-gimg} & RGB & \bf 1.0000 & 0.6382& 0.5390& 0.3154& 0.7663& 0.7677& 0.7406& 0.8599& 0.9374& 0.0000& 0.0489\\
    Proposed (single) & Spec.& 0.9652 &\bf 0.7704 &\bf 0.6063&\bf 0.8487& 0.7446& 0.6675& 0.7276& 0.7672& 0.6226& \bf 0.4014&\bf 0.1585\\
    Proposed (ensemble) & RGB+Spec. & \bf 1.0000& 0.6839& 0.5292& 0.6831&\bf 0.8180&\bf 0.7695&\bf 0.7712&\bf 0.8976& \bf 0.9379& 0.0000 & 0.0369\\
    \end{tabular}%
        }
    \end{center}
\end{table*}
\begin{table*}[t]
    \caption{Experiment result under 11 models (AP)}
    \label{tb:result_ap}
    \begin{center}
        \scalebox{0.80}{
    \begin{tabular}{c|c||c|c|c|c|c|c|c|c|c|c|c|c}
    Methods& Input image & PGGAN& StyleGAN & StyleGAN2 & BigGAN & CycleGAN & StarGAN & GauGAN & CRN & IMLE & SAN & Deepfake& mean\\ \hline
    \hline
    Wang's method\cite{CNN-gimg} & RGB & \bf 100.00 & 98.51& 97.99& 88.23& 96.82& 95.45& 98.09& 98.95& 99.42& 63.88& \bf 66.27& 91.24\\
    Proposed (single) & Spec.& 99.56 & 90.92 & 86.38& 94.27& 92.30& 82.22& 84.75& 82.90& 62.29& 73.91& 60.13& 82.69\\
    Proposed (ensemble) & RGB+Spec. & \bf 100.00& \bf 98.76& \bf 98.02& \bf 94.37& \bf 98.17& \bf 95.53& \bf 98.50& \bf 99.07& \bf 99.52& \bf 69.24& 66.21& \bf 92.51\\
    \end{tabular}%
        }
    \end{center}
\end{table*}

\section{Conclusion}
We proposed a universal detector with enhanced spectrums for detecting CNN-generated images. The proposed ensemble was confirmed to outperform the state-of-the-art under the use of 11 models, where the classifier for the detection was trained by using image generated only from one model, PGGAN.


\end{document}